\documentclass[12pt]{article}
\usepackage[]{graphics}
\usepackage{amsmath, amssymb, amsthm}
\usepackage{authblk}
\usepackage{epsfig}
\usepackage{verbatim}
\usepackage{algorithmic}
\usepackage{setspace}
\usepackage{xcolor}
\usepackage[font=footnotesize]{caption}

\newtheorem{theorem}{Theorem}

\newtheorem{lemma}{Lemma}
\newenvironment{changemargin}{%
\begin{list}{}{%
\setlength{\topsep}{0pt}%
\setlength{\leftmargin}{0.5cm}%
\setlength{\rightmargin}{0.5cm}%
\setlength{\listparindent}{\parindent}%
\setlength{\itemindent}{\parindent}%
\setlength{\parsep}{\parskip}%
}%
\item[]}{\end{list}}
\newtheoremstyle{appendixthm}
{\topsep}
{\topsep}%
{\em}
{}
{\bfseries}
{}
{0.18cm}
{\thmname{#1}\thmnumber{ #2}\thmnote{ #3.}}
\theoremstyle{appendixthm}

\newtheoremstyle{example}
{\topsep}
{0.8cm}%
{}
{}
{\bfseries}
{.}
{0.3cm}
{\thmname{#1}\thmnumber{ #2}\thmnote{ #3}}
\theoremstyle{example}
\newtheorem{example}{Example}[section]

\newcommand{\bexample}{\begin{changemargin}\begin{example}}
\newcommand{\eexample}{\end{example}\end{changemargin} ~\\ \vspace{0.3cm}}

\renewcommand{\qed}{\hfill $\Box$ \\ \vspace{0.3cm}}
\renewcommand{\proof}{\noindent {\bf Proof: }}

\newcommand\argmax{\mathop{\mbox{{\rm argmax}}}\limits}

\newcommand\Fscr{\mathcal{F}}
\newcommand\Xscr{\mathcal{X}}

\title{Comments on \\ the Du-Kakade-Wang-Yang Lower Bounds}

\author[1,2]{Benjamin Van Roy}
\author[1]{Shi Dong}
\affil[1]{Stanford University}
\affil[2]{DeepMind}

\begin{document}

\maketitle

\abstract{Du, Kakade, Wang, and Yang \cite{Du2019bound} recently established intriguing lower bounds on sample complexity, which suggest that reinforcement learning with a misspecified representation is intractable.  Another line of work, which centers around a statistic called the {\it eluder dimension} \cite{russo2014posterior,russo2013eluder}, establishes tractability of problems similar to those considered in \cite{Du2019bound}.  We compare these results and reconcile interpretations.}

\section{Introduction}

Du, Kakade, Wang, and Yang \cite{Du2019bound} recently established intriguing lower bounds on the sample complexity of reinforcement learning with a misspecified representation.  Versions of the lower bound apply to model learning, value function learning, and policy learning.  The cornerstone of their analysis is a basic problem, embedded in each of their results, of bandit learning with a misspecified linear model.  The problem is one of finding a needle in a haystack: an agent must identify among an exponentially large number of actions the only one that generates rewards.  This obviously requires exponentially many trials.  One might hope that with a suitable choice of features, by using a linearly parameterized approximation to generalize across actions, the agent can efficiently identify the rewarding action.  However, as established in \cite{Du2019bound}, even if the linear model can approximate rewards with uniform accuracy across actions, an exponentially large number of trials may be required.

Another line of work, which centers around a statistic called the {\it eluder dimension} \cite{russo2014posterior,russo2013eluder}, offers additional insight into bandit learning.
In particular, an analysis from \cite{russo2014posterior,russo2013eluder} suggests qualitatively different behavior, indicating that if the linear model can approximate rewards uniformly with sufficient accuracy, the agent {\it can} efficiently identify the rewarding action.  In this technical note, we reconcile what may appear to be  contradictory narratives stemming from these two lines of analysis.

What we find is that the example used to establish the lower bound of \cite{Du2019bound} violates assumptions imposed by the upper bound of \cite{russo2014posterior,russo2013eluder}.  In essence, the latter requires features to be sufficiently informative.  The example that establishes the lower bound makes use of features that are uninformative though they enable accurate approximation of rewards, in some sense.  Upon sharing an early version of this technical note, we discovered that Lattimore and Szepesv\'{a}ri also arrived at a similar conclusion and are working toward a deeper analysis of this issue \cite{Lattimore2019}.

We begin by formulating in the next section a class of bandit learning problems.  Then, in Section \ref{se:lower}, we discuss the special case of finding a needle in a haystack and a lower bound that can be established via the analysis of \cite{Du2019bound}.  We next establish an upper bound based on arguments developed in \cite{russo2014posterior,russo2013eluder}.
Finally, we interpret these results in a manner that reconciles narratives.

\section{A Bandit Learning Problem}
\label{se:bandit}

Consider a bandit learning problem characterized by a pair $(\Xscr, \Fscr)$, where
$\Xscr$ is a non-singleton finite set and $\Fscr$ is a class of reward functions that each maps $\Xscr$ to $[0,1]$.
Let $f^* \in \Fscr$ denote the reward function that generates observed outcomes.
An agent begins with knowledge of $(\Xscr, \Fscr)$ but not $f^*$.  The agent
operates over time periods $t=0,1,2,\ldots$, in each period
selecting an action $x_t \in \Xscr$ and observing a deterministic outcome $y_{t+1} = f^*(x_t)$. 

Suppose that before making its first decision, the 
agent is provided with a feature map $\phi:\Xscr \mapsto \Re^d$, which assigns a 
feature vector $\phi(x) \in \Re^d$ to each action $x \in \Xscr$.  Let $\tilde{f}_\theta(x) = \theta^\top \phi(x)$ 
denote a linear combination of features with coefficients $\theta \in \Re^d$.
Suppose the agent is also informed that $f^*$ can be closely approximated by a linear combination
of features in the sense that
\begin{equation}
\label{eq:accuracy}
\min_{\theta \in \Re^d : \|\theta\|_2 \leq 1} \|f^* - \tilde{f}_\theta\|_\infty \leq \epsilon,
\end{equation}
for some known $\epsilon > 0$.

We consider assessing an agent based on the expected number of trials it requires to identify an
$\epsilon'$-optimal action, for some tolerance parameter $\epsilon' > 0$. Here we define an action $x$ to be $\epsilon'$-optimal if 
\[
	f^*(x) \geq \max_{x' \in \Xscr} f^*(x') -  \epsilon'.
\]
The agent's algorithm takes $(\Xscr, \Fscr)$, $\phi$, $\epsilon$, and $\epsilon'$ as input.
The expectation is over algorithmic randomness in the event that the agent uses a randomized algorithm.

\section{A Lower Bound}
\label{se:lower}

The analysis of \cite{Du2019bound} yields the following lower bound.
\begin{theorem}
\label{th:lower}
For all learning algorithms, $\epsilon > 0$,  and $\epsilon' \in [0,0.5)$, for $d \rightarrow \infty$, 
there exists $(\Xscr, \Fscr)$, $f^* \in \Fscr$, and a feature map $\phi:\Xscr \mapsto \Re^d$ satisfying 
$$\min_{\theta \in \Re^d : \|\theta\|_2 \leq 1} \|f^* - \tilde{f}_{\theta}\|_\infty \leq \epsilon,$$
such that the expected number of trials required to identify an $\epsilon'$-optimal action is $\Omega(2^d)$.
\end{theorem}
\noindent The expectation is over algorithmic randomness, in the event that the agent employs a randomized algorithm.  This result indicates that an exponentially large number of trials can be required even if the agent knows features that can accurately approximate rewards.  As demonstrated in \cite{Du2019bound}, this can be established via a simple example which we will now discuss.

Consider a function class $\Fscr$ comprised of one-hot functions.  In particular, $|\Fscr| = |\Xscr|$ and, for each $x \in \Xscr$, there is a function
$f \in \Fscr$ for which $f(x) = 1$ and $f(y) = 0$ for all $y \neq x$.  Let $f^* \in \Fscr$ denote the unknown function of interest
and $x^* \in \Xscr$ be such that $f(x^*) = 1$.  To produce coefficients 
$\theta_t$ such that
$$\|f^* - \tilde{f}_{\theta_t}\|_\infty < 0.5,$$
the agent must identify $x^*$.  It is easy to see that this requires $\Omega(2^d)$ trials.

Now suppose the agent knows features that can accurately approximate rewards, in the sense of (\ref{eq:accuracy}).
Lemma 5.1 of \cite{Du2019bound}, restated here, allows us to select features that are uninformative while meeting such an accuracy requirement.
\begin{lemma}
\label{le:random-features}
For all non-singleton finite $\Xscr$, $\epsilon > 0$, and $d \geq 8 \ln (|\Xscr|) / \epsilon^2$,
there exists $\phi:\Xscr\mapsto \Re^d$ such that, for all $x,y \in \Xscr$ with $x\neq y$, $\|\phi(x)\|_2 =1$ and
$|\phi^\top(x) \phi(y)| \leq \epsilon$.
\end{lemma}
\noindent Fixing $\epsilon > 0$ and letting $d \geq 8 \ln (|\Xscr|) / \epsilon^2$, this lemma 
prescribes a feature map $\phi$ such that
\begin{eqnarray*}
\max_{f \in \Fscr} \min_{\theta \in \Re^d : \|\theta\|_2 \leq 1} \|f - \tilde{f}_\theta\|_\infty 
&=& \max_{f \in \Fscr} \min_{\theta \in \Re^d : \|\theta\|_2 \leq 1} \max_{x \in \Xscr} |f(x) - \theta^\top \phi(x)| \\
&\leq& \max_{z \in \Xscr} \min_{y \in \Xscr} \max_{x \in \Xscr} |{\bf 1}(x=z) - \phi^\top(y) \phi(x)|  \\
&\leq& \epsilon.
\end{eqnarray*}
Since this feature map $\phi$ does not depend on $f^*$, it does not offer any information that assists
in identifying $x^*$.  As such, given these features, the agent still requires $\Omega(2^d)$ trials.

\section{An Upper Bound}

The following result offers an upper bound for an agent that  selects actions that aim to quickly hone in on $f^*$.  The result
is general, applying not only to the ``needle in a haystack'' instance in Section \ref{se:lower} but more broadly to the bandit learning problem in Section \ref{se:bandit}.
\begin{theorem}
\label{th:upper}
For all $\epsilon, \epsilon' > 0$, $(\Xscr, \Fscr)$, $f^* \in \Fscr$, feature maps
$\phi:\Xscr\mapsto \Re^d$ such that
$$\min_{\theta \in \Re^d : \|\theta\|_2 \leq 1} \|f^* - \tilde{f}_{\theta}\|_\infty \leq \epsilon,$$
and
\begin{equation}
	\label{eq:upper-bound-condition}
	\epsilon' \geq 2 \epsilon \left(1 + \sqrt{3d \log\left(1 + \frac{1}{d \epsilon^2}\right)}\right),
\end{equation}
there exists a learning algorithm that identifies an $\epsilon'$-optimal action within $3d \log(1 + 1 / (d \epsilon^2))$ trials.
\end{theorem}
\noindent The theorem can be established via an analysis developed in \cite{russo2014posterior,russo2013eluder} to bound
the eluder dimension of linear function classes.  For convenience, we provide a self-contained proof in the appendix,
which adapts those provided in the papers.

Theorem \ref{th:upper} establishes a sense in which the agent learns efficiently, so long as \eqref{eq:upper-bound-condition} is satisfied.  
Our discussion in the next section offers some intuition motivating this constraint on $\epsilon$ and $d$, and 
aims to reconcile the efficiency result with Theorem \ref{th:lower}.

\section{Discussion}

The lower bound established by Theorem \ref{th:lower} suggests that an accurate linear representation does not
suffice for efficient learning while the upper bound established by Theorem \ref{th:upper} suggests it does.
Reconciling the results requires careful examination how examples that establish the lower bound 
violate assumptions under which the upper bound holds.
Examples that establish the lower bound involve features of the kind identified by Lemma \ref{le:random-features}.
The constraint on dimension required by this lemma can be written as
\begin{equation}
\label{as:lower}
\epsilon \sqrt{d} \geq \sqrt{8 \ln (|\Xscr|)}.
\end{equation}
Letting $\epsilon' = 1/4$ to simplify the comparison, 
the requirement \eqref{eq:upper-bound-condition} of the upper bound established by Theorem \ref{th:upper} is satisfied if 
\begin{equation}
\label{as:upper}
\epsilon \sqrt{d} \leq \frac{1}{100}.
\end{equation}
Hence, the upper bound holds when $\epsilon \sqrt{d}$ is small while the lower bound holds when $\epsilon \sqrt{d}$ is large.  These constraints can be viewed as complementarity conditions, requiring $\epsilon$ and $d$ to suitably offset one another.

\begin{figure}[h]
\centering
\captionsetup{width=0.8\linewidth}
\includegraphics[width=6.5cm]{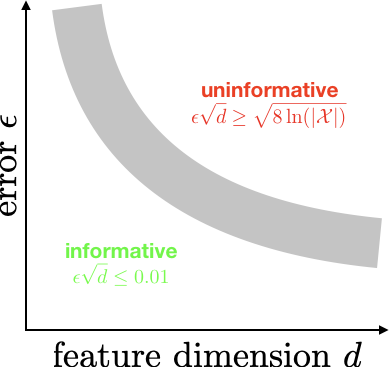}
\caption{\footnotesize Regimes under which the lower and upper bounds are satisfied.  The lower bound is satisfied when features are uninformative.  The upper bound is satisfied when features are sufficiently informative.}
\label{fig:regimes}
\end{figure}

Recall that $d$ is the number of features while $\epsilon$ is the error within which they can approximate $f^*$.  When $\epsilon \sqrt{d}$ is large, the error is large relative to the number of features, or the number of features is large relative to the error, or both are large.  The proof of the lower bound is constructive, and involves identifying features that achieve a particular level of error.  These features can be generated without any information about $f^*$, so they must not be helpful in learning $f^*$.  As such, (\ref{as:lower}) captures levels of error that can be achieved when features offer no useful information.  Clearly, as the number of features increases, even if they are uninformative, error should decrease.  So we could also view this result as capturing a rate at which error can decrease as uninformative features are incorporated.

If we apply the upper bound to the hard instance of finding a needle in a haystack, the fact that the result guarantees efficient learning implies that the features must be required to offer useful information and must therefore depend on $f^*$.  To ensure this, the error needs to be small relative to the number of features, or the number of features needs to be small relative to the error.  This is intuitive: if few features lead to small error, the features must be informative.  Figure \ref{fig:regimes} illustrates how the lower and upper bounds reflect different regimes in the space of $(\epsilon, d)$ pairs.  The grey region represents pairs that satisfy neither (\ref{as:lower}) nor (\ref{as:upper}).  The upper bound of Theorem \ref{th:upper}
should apply within some of this grey region, as the constraint $\epsilon \sqrt{d} \leq 0.01$ is much stronger than and chosen to simplify (\ref{eq:upper-bound-condition}).

Note that the requirement (\ref{as:lower}) for the lower bound depends on the number of actions $|\Xscr|$.  This is because, as the number of actions grows, the number of uninformative features required to achieve error $\epsilon$ also grows.  On the other hand, the requirement (\ref{as:upper}) does not exhibit any dependence on the number of actions.  As $|\Xscr|$ increases, the uninformative regime identified by the lower bound shrinks, and the grey region of Figure \ref{fig:regimes} grows.

\bibliographystyle{unsrt} 
\bibliography{bibliography}

\appendix

\section{Proof of Theorem \ref{th:upper}}

We will establish the result for an algorithm that selects actions according to
$$x_t \in \argmax_{x \in \Xscr} \left(\max_{\theta \in \Theta_t} \tilde{f}_{\theta}(x) - \min_{\theta \in \Theta_t} \tilde{f}_{\theta}(x)\right),$$
where
$$\Theta_t = \left\{\theta \in \Re^d: \|\theta\|_2 \leq 1, \left(\sum_{\tau=0}^{t-1} (y_{\tau+1} - \tilde{f}_\theta(x_\tau))^2\right)^{1/2} \leq \epsilon \sqrt{t} \right\},$$
Note that the set $\Theta_t$ is nonempty because $\theta^* \in \Theta_t$, since $|y_{t+1} - \tilde{f}_{\theta^*}(x_t)| \leq \epsilon$.

Let
$$w_t = \max_{\theta \in \Theta_t} \tilde{f}_{\theta}(x_t) - \min_{\theta \in \Theta_t} \tilde{f}_{\theta}(x_t),$$
To prove the result, we first bound the number of times $w_t$ can be larger than $2 \epsilon \sqrt{t}$.  
\begin{lemma}
	\label{le:width-bound}
	If $w_\tau \geq 2 \epsilon \sqrt{\tau}$, for $\tau=0,\ldots,t-1$, then
	$$t \leq 3d \log\left(1 + \frac{1}{d \epsilon^2}\right).$$
\end{lemma}
\proof Let $w_\tau \geq 2 \epsilon \sqrt{\tau}$ for $\tau=0,\ldots,t-1$.
For shorthand, let $\phi_\tau = \phi(x_\tau)$, $\Phi_\tau = \sum_{k=0}^{\tau-1} \phi_k \phi_k^\top$, and $\Psi_\tau = \Phi_\tau + \epsilon^2 \tau I$.
Let
$$\tilde{\Theta}_\tau = \left\{\rho \in \Re^d: \|\rho\|_2 \leq 2, \left(\sum_{k=0}^{\tau-1}\big( \rho^\top \phi_k(x_k)\big)^2\right)^{\frac{1}{2}} \leq 2 \epsilon \sqrt{\tau} \right\},$$
and note that, for all $\theta, \theta' \in \Theta_\tau$, we have $\theta-\theta' \in \tilde{\Theta}_\tau$.
Since $\rho^\top \Phi_\tau \rho = \sum_{k=0}^{\tau-1} (\rho^\top \phi_k)^2$,
\begin{eqnarray*}
	\tilde{\Theta}_\tau
	&=& \left\{\rho \in \Re^d: \|\rho\|_2 \leq 2, \rho^\top \Phi_\tau \rho \leq 4 \epsilon^2 \tau \right\} \\
	&=& \left\{\rho \in \Re^d: \rho^\top ( \epsilon^2 \tau I) \rho \leq 4 \epsilon^2 \tau, \rho^\top \Phi_\tau \rho \leq 4 \epsilon^2 \tau \right\} \\
	&\subseteq& \left\{\rho \in \Re^d: \rho^\top \Psi_\tau \rho \leq 8 \epsilon^2 \tau \right\}.
\end{eqnarray*}
Hence, 
\begin{eqnarray}
	w_\tau
	&=& \max_{\theta \in \Theta_\tau} \tilde{f}_\theta(x_\tau) - \min_{\theta \in \Theta_\tau} \tilde{f}_{\theta}(x_\tau) \nonumber\\
	&=& \max_{\theta, \theta' \in \Theta_\tau} (\theta - \theta')^\top \phi_\tau \nonumber\\
	&\leq& \max_{\rho \in \tilde{\Theta}_\tau} \rho^\top \phi_\tau \nonumber\\
	&\leq& \sup_{\rho: \rho^\top \Psi_\tau \rho \leq 8 \epsilon^2 \tau} \rho^\top \phi_\tau \nonumber\\
	&=& \sqrt{8 \epsilon^2 \tau \phi_\tau^\top \Psi_\tau^{-1} \phi_\tau} \label{eq:w-upper}.
\end{eqnarray}
Combining \eqref{eq:w-upper} and the fact that $w_\tau \geq 2\epsilon\sqrt{\tau}$, we have that $\phi_\tau^\top \Psi_\tau^{-1} \phi_\tau \geq \frac{1}{2}$.

Note that $\Psi_\tau = \Psi_{\tau-1} + \phi_{\tau-1} \phi_{\tau-1}^\top$.  
Let $\lambda = \epsilon^2 t$.
The Matrix Determinant Lemma yields
\begin{eqnarray*}
	\det \Psi_t 
	&=&  (1 + \phi_{t-1}^\top \Psi_{t-1}^{-1} \phi_{t-1}) \det \Psi_{t-1} \\
	&\geq& \frac{3}{2} \det \Psi_{t-1} \geq \cdots \\
	&\geq& \left(\frac{3}{2}\right)^t \det (\lambda I)  = \left(\frac{3}{2}\right)^t \lambda^d.
\end{eqnarray*}
The determinant of a positive semidefinite matrix is the product of the eigenvalues, whereas the trace is their sum.  
As such, the inequality of arithmetic and geometric means yields
\begin{eqnarray*}
\det \Psi_t 
&\leq& \left(\frac{{\rm trace}(\Psi_t)}{d}\right)^d \\
&=& \left(\frac{{\rm trace}(\lambda I) + \sum_{\tau=0}^{t-1} {\rm trace}(\phi_\tau \phi_{\tau}^\top)}{d}\right)^d \\
&\leq& \left(\lambda + \frac{t}{d}\right)^d.
\end{eqnarray*}
It follows that
$(3/2)^t \lambda^d \leq (\lambda + t/d)^d,$
and therefore,
\[
	t \leq d \log_{\frac{3}{2}}\left(1 + \frac{t}{\lambda d}\right) \leq 3 d \log\left(1 + \frac{1}{d \epsilon^2}\right),
\]
as desired.
\qed

Note that
\begin{eqnarray*}
\max_{x \in \Xscr} f^*(x') - f^*(x)
&\leq& \max_{x' \in \Xscr} \tilde{f}_{\theta^*}(x') - \tilde{f}_{\theta^*}(x) + 2 \epsilon \\
&\leq& \max_{\theta \in \Theta_t} \tilde{f}_\theta(x) - \min_{\theta \in \Theta_t} \tilde{f}_{\theta}(x) + 2 \epsilon.
\end{eqnarray*}
Hence, when $\max_{\theta \in \Theta_t} \tilde{f}_\theta(x) - \min_{\theta \in \Theta_t} \tilde{f}_{\theta}(x) \leq \epsilon' - 2 \epsilon$,
for any action $x$, that action has been identified as an $\epsilon'$-optimal action.
It follows that action $x_t$ has been identifies as $\epsilon'$-optimal if $w_t \leq \epsilon' - 2\epsilon$.

Recall that
$$\epsilon' \geq 2 \epsilon \left(1 + \sqrt{3d \log\left(1 + \frac{1}{d \epsilon^2}\right)}\right)$$
By Lemma \ref{le:width-bound}, if $w_\tau \geq \epsilon' \geq 2 \epsilon \sqrt{\tau}$, for $\tau=0,\ldots,t-1$, 
then
$$t \leq 3d \log\left(1 + \frac{1}{d \epsilon^2}\right).$$
This inequality implies that $\epsilon' \geq 2 \epsilon \sqrt{t}$.
It follows that an $\epsilon'$-optimal action is identified within $3d \log(1 + 1 / d \epsilon^2)$ trials.

\end{document}